\documentclass[runningheads,a4paper]{llncs}

\usepackage{amssymb}
\usepackage[ruled,algonl]{algorithm2e}
\usepackage[latin1]{inputenc}
\usepackage{amsmath}
\usepackage{amsfonts}
\usepackage{float}
\usepackage{tikz}
\usepackage{amssymb}
\usepackage{graphicx,graphics}

\usepackage{url}
\urldef{\mailsc}\path|naresh,sastry@ee.iisc.ernet.in|
\newcommand{\keywords}[1]{\par\addvspace\baselineskip
\noindent\keywordname\enspace\ignorespaces#1}

\begin{document}

\mainmatter  

\title{Polyceptron: A Polyhedral Learning Algorithm}

\titlerunning{Polyceptron}

%
%
\author{}
%
\authorrunning{}

\institute{
}

\toctitle{Lecture Notes in Computer Science}
\tocauthor{Authors' Instructions}
\maketitle
\def \bw {\tilde{\mathbf{w}}}
\def \by {\tilde{\mathbf{y}}}
\def \bx {\tilde{\mathbf{x}}}
\def \bz {\tilde{\mathbf{Z}}}
\def \bX {\tilde{\mathbf{X}}}
\def \xx {\mathbf{x}}
\def \yy {\mathbf{y}}
\def \ee {\mathbf{e}}
\def \ww {\mathbf{w}}
\def \zz {\mathbf{z}}
\def \amax {\operatorname{argmax}}
\def \amin {\operatorname{argmin}}

\begin{abstract}
In this paper we propose a new algorithm for learning polyhedral classifiers which we call as {\em Polyceptron}.
It is a Perceptron like algorithm which updates the parameters only when the current classifier misclassifies any training data.
We give both batch and online version of Polyceptron algorithm. Finally we give experimental results to show the effectiveness of our approach.
\end{abstract}

\section{Introduction}\label{Sec:Introduction}
A polyhedral set is a convex set formed by intersection of finite collection of closed half spaces \cite{Rockafellar}.
Many interesting properties of polyhedral sets make them useful in many fields. An important property of polyhedral sets is that they can be used to approximate any convex subset of $\Re^d$.
This property of polyhedral sets makes learning of polyhedral regions an interesting problem in pattern recognition. Many binary classification
problems are such that all the class $C_1$ examples are concentrated in a single convex region with the class $C_2$ examples being all around that region.
Then the region of class $C_1$ can be well captured by a polyhedral set.

One possible way of learning a classifier in this case is to
fit a minimum enclosing hypersphere
in the feature space to include most of the class $C_1$ examples inside the hypersphere \cite{Tax1999} and all the class $C_2$ examples are considered as outliers.
This problem is formulated in a large margin one-class classification framework, which is a variant of the well known Support Vector Machine (SVM) method
\cite{Burges1998}. In such techniques, the nonlinearity in the data is captured simply by choosing an appropriate kernel function.
Although the SVM methods often give good classifiers, with a non-linear kernel function, the final classifier may not provide good geometric insight on the
local behavior of the classifier in original feature space.

Another well known approach to learn polyhedral sets is the top-down decision tree method. In a binary classification problem, a top-down decision tree represents
each class region as a union of polyhedral sets \cite{Breiman1984}. When all positive examples belong to a single polyhedral set, one
can expect a decision tree learning algorithm to learn a tree where each non-leaf node has
one of the children as a leaf (representing negative class)
and there is only one path leading to a leaf for
the positive class. Such a decision tree (which is also called a decision list) would represent the
polyhedral set exactly. However, generic top-down decision tree algorithms fail to learn a single polyhedral set well.

As opposed to such general purpose methods, there are many fixed structure approaches proposed for learning polyhedral classifiers. In case of polyhedral classifiers, the structure can be fixed by
fixing the number of hyperplanes. We discuss these fixed structure approaches in the next section after describing the problem of learning polyhedral classifier.

In this paper we propose a Perceptron like algorithm to learn polyhedral classifier which we call {\em Polyceptron}. The Polyceptron algorithm is based on minimization of Polyceptron criterion
which is designed in the same spirit as the Perceptron criterion.
In this paper, we propose both batch and online version of the Polyceptron algorithm.

The rest of the chapter is organized as follows. In Section \ref{Sec:polyhedral classifier}
we formulate polyhedral learning problem and discuss the credit assignment problem corresponding to polyhedral classifier learning and how different approaches address
this problem. We describe the Polyceptron algorithm in section \ref{Sec:Polycep}.
Experimental results are given in Section \ref{Sec:Experiments}. We conclude the paper in Section \ref{Sec:Conclusion}.

\section{Polyhedral Classifier}\label{Sec:polyhedral classifier}
Let $S=\{(\xx_1 , y_1),\ldots,(\xx_N,y_N)\}$ be the training dataset,
where $(\xx_n,y_n)\in \Re^d\times\{-1,+1\},\forall n$.
Let $C_1$ (positive class) be the set of
points for which $y_n=1$ and let $C_2$ (negative class) be the set of points for which $y_n=-1$.

\subsection{Polyhedral Separability}
\begin{figure}
 \begin{center}
\scalebox{.35}{\input{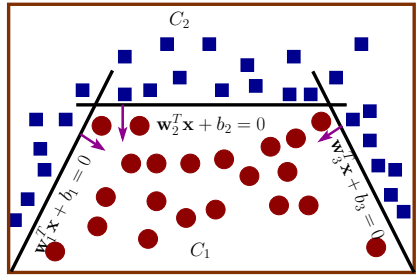}}
 \end{center}
\caption{An example of polyhedrally separable sets $C_1$ and $C_2$. Arrows are pointing towards positive side of hyperplanes.}
\label{Fig:poly_sep_example}
\end{figure}

Two sets $C_1$ and $C_2$ in $\Re^d$ are $K$-polyhedral separable \cite{Astorino} if there exists a set
of $K$ hyperplanes having parameters, $(\ww_k,b_k),\;k=1\ldots K$, with $\ww_k \in  \Re^d,\;b_k \in \Re,\; k=1\ldots K$, such that
\begin{enumerate}
 \item $\forall \; \xx \in C_1,\;\forall\; k\in\{1,\ldots, K\},\; \ww_k^T\xx+b_k \geq 0$, and
\item $\forall \; \xx \in C_2,\;\exists\text{ at least one }k\in \{1,\ldots, K\},\;\text{s.t.}\;\ww_k^T\xx+b_k < 0$
\end{enumerate}
Thus, two sets $C_1$ and $C_2$ are $K$-polyhedral separable if $C_1$ is contained in a convex polyhedral
set formed by intersection of $K$ half spaces and the points of set $C_2$ are outside this polyhedral set.
Fig.~\ref{Fig:poly_sep_example} shows an example of sets $C_1$ and $C_2$ which are polyhedrally separable.
\subsection{Polyhedral Classifier}
Consider the problem of learning a polyhedral set with $K$, the number of hyperplanes needed, known. Given the parameters of the $K$ hyperplanes,
$\ww_k \in \Re^d,b_k \in \Re,\;k=1\ldots K$, define a function $h$ by
\begin{equation}\label{eq:modelfunction}
  h(\xx,\Theta)=\min_{k\in \{1,\ldots, K\}}(\ww_k^T\xx+b_k)
\end{equation}
where $\Theta=\{(\ww_1,b_1),\ldots,(\ww_K,b_K)\}$. Clearly if $h(\xx,\Theta)\geq 0$, then the condition $\ww_k^T\xx+b_k \geq 0$ is satisfied for all $k\in\{1,\ldots, K\}$ and the point $\xx$ can
be assigned to set $C_1$. Similarly if $h(\xx,\Theta)< 0$, there exists at least one $k\in\{1,\ldots,K\}$, for which $\ww_k^T\xx+b_k <0$ and the point $\xx$
can be assigned to set $C_2$. Thus, the polyhedral classifier can be expressed as
\begin{equation}\label{eq:poly-classifier}
 f(\xx,\Theta)=\text{sign}(h(\xx,\Theta))=\text{sign}\big{(}\min_{k\in \{1,\ldots, K\}}(\ww_k^T\xx+b_k)\big{)}
\end{equation}
Let $\bw_k =[\ww_k~b_k]^T \in \Re^{d+1}$ and let $\bx_n=[\xx_n~1]^T \in \Re^{d+1}$. We
now express the earlier inequalities as $\bw_k^T\bx>0$ and so on.

\subsection{Credit Assignment Problem}\label{subsec:credit}
Learning a polyhedral classifier with $K$ hyperplanes effectively solving $K$ linear classification
subproblems. The key issue in polyhedral learning is identifying
the subproblems. Assume that the data is polyhedrally separable. Let the number of hyperplanes required to form a polyhedral classifier be $K$. By the definition of polyhedral separability, it is clear that
all points in the set $C_1$ should fall on the positive side of all the hyperplanes.
In other words, all points in set $C_1$ forms one class in every subproblem.
On the other hand, for every hyperplane, there exists a subset of points of set $C_2$ which falls on its negative side. Which means that for each subproblem there is a subset of points of set $C_2$
that forms the other class.
\begin{figure}[t]
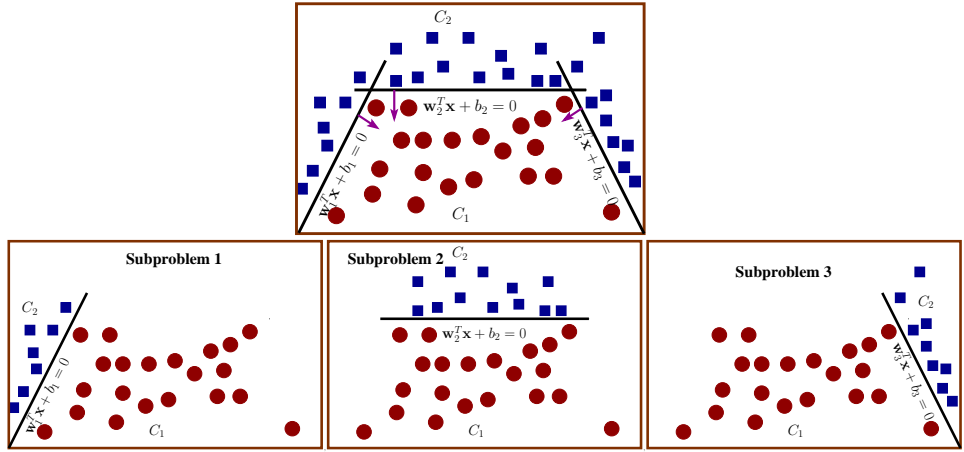

\begin{center}
 \scalebox{.3}{\input{poly_sep.pstex_t}}  \\
\scalebox{.27}{\input{poly_sep1.pstex_t},\input{poly_sep2.pstex_t},\input{poly_sep3.pstex_t}}
\caption{A polyhedral classification problem and its three subproblems.
Every subproblem boils down to a linear classification problem.}
\label{Fig:credit-assign}
\end{center}
\end{figure}
Thus the goal is to cluster or divide the points in set $C_2$ into $K$ disjoint subsets
such that each subset is linearly separable with points in set $C_1$.

Fig.~\ref{Fig:credit-assign} shows an example of polyhedral classification
problem and the corresponding three subproblems. Thus learning a polyhedral set, in a sense,
is equivalent to clustering because after such clustering, the $K$ subproblems are easy to
solve. However, this clustering problem is hard because the cluster membership of any point is actually determined by the underlying polyhedral set which we do not know and are trying to learn. Hence one has to keep guessing different ways of splitting points of class $C_2$. Clearly the problem is combinatorial and is shown to be NP-complete \cite{Megiddo}.

There is another way to look at this problem. If for each point $\xx_n\in C_2$, we are given for which $k$ we have $\ww_k^T\xx_n+b_k<0$, then we know the cluster membership of $\xx_n$ and hence would know which subproblem $\xx_n$ should be in. Thus what we do not know is when $h(\xx_n)$ in equation (\ref{eq:modelfunction}) is negative, which $k$ is responsible for it. Hence this is also termed as credit assignment problem \cite{Megiddo}.
Any polyhedral classifier learning algorithm has to devise a mechanism for efficiently handling this credit assignment problem.

\subsection{Current Approaches to Learn Polyhedral Classifier}
One can handle the underlying clustering problem by solving constrained optimization problems \cite{Astorino,Carlotta,Murat2008}.
These optimization problems minimize the classification errors subject to the separability conditions. Note that these optimization problems are non-convex
even though we are learning a convex set. Here all the positive examples satisfy each of a given set of linear inequalities (that defines the half spaces
whose intersection is the polyhedral set). However, each of the negative examples fail to satisfy one (or more) of these
inequalities and we do not know a priori which inequality each negative example fails to satisfy. Thus constraint on each of the negative examples is
logical `or' of the linear constraints which makes the optimization problem non-convex.

In \cite{Astorino}, this problem is solved by first enumerating all possibilities for misclassified negative examples (e.g., which hyperplanes
is responsible for a negative example to get misclassified and for each negative example there could be many such hyperplanes) and then solving a linear program
for each possibility to find descent direction. This approach becomes computationally very expensive.

If, for every point falling outside the polyhedral set, it is known beforehand which of the linear inequalities it will satisfy (in other words,
negative examples
for each of the linear subproblems are known), then the problem becomes much easier. In that case, the problem becomes one
of solving $K$ linear classification problems independently. But this assumption is very unrealistic. Polyhedral learning approach in \cite{Murat2008} assumes that for every linear subproblem, a small subset of negative examples is known and proposes a cyclic
optimization algorithm. Still, their assumption of knowing subset of negative examples corresponding to
every linear subproblem is not realistic in many practical applications.

Recently, a probabilistic discriminative model has been proposed in \cite{Naresh10} using logistic function to learn a polyhedral classifier.
It is an unconstrained framework and a simple expectation maximization algorithm is employed to learn the parameters. However, the approach proposed
in \cite{Naresh10} is a batch algorithm and there is no incremental variant of this algorithm.

\section{Polyceptron}\label{Sec:Polycep}
Here we propose a Perceptron like algorithm for learning polyhedral classifier which we call {\em Polyceptron}.
The goal of Polyceptron is to find the parameter set $\Theta=\{\bw_1,\ldots,\bw_K\}$ of $K$ hyperplanes such that point $\xx_n \in C_1$ will have
$h(\xx_n,\Theta)=\min_{\substack{k\in \{1,\ldots, K\}}}(\bw_k^T\bx_n) >0$, whereas point $\xx_n \in C_2$ will have
$h(\xx_n,\Theta)=\min_{\substack{k\in \{1,\ldots, K\}}}(\bw_k^T\bx_n)<0$. Since $y_n \in \{-1,1\}$ is the class label
for $\xx_n$, we want that each point $\xx_n$ should satisfy $y_nh(\xx_n,\Theta) >0$.

Polyceptron algorithm finds polyhedral classifier by minimizing the {\em Polyceptron criterion} which is defined as follows.
\begin{eqnarray}\label{eq:poly-crit}
  E_P(\Theta)&:=&-\sum_{n=1}^ny_n h(\xx_n,\Theta)I_{\{y_nh(\xx_n,\Theta)<0\}}
\end{eqnarray}
Where $I_{\{A\}}$ is an indicator function which takes value 1 if its argument A is true and 0 otherwise.
Polyceptron criterion assigns zero error for a correctly classified point. On the other hand, if a point $\xx_n$ is misclassified,
the Polyceptron criterion assigns error of $-y_nh(\xx_n,\Theta)$.

\subsection{Batch Polyceptron}
Batch Polyceptron minimizes the Polyceptron criterion $E_P(\Theta)$ as defined in equation~\ref{eq:poly-crit}, considering all the data points at a time.
Batch Polyceptron works in the following way.
Given parameters of $K$ hyperplanes, $\bw_1 \ldots \bw_K$, define sets $S_k=\{\xx_n|\bx_n^T\bw_k \leq \bx_n^T\bw_j,~\forall j \neq k\}$
where we break ties by putting $\xx_n$ in the set $S_k$ with least $k$ if $\bx_n^T\bw_k \leq \bx_n^T\bw_j \forall j \neq k$ is satisfied by more than
one $k\in\{1,\ldots,K\}$. The sets $S_k$ are disjoint. We can now write $E_P(\Theta)$ as
\begin{equation}\label{error2}
 E_P(\Theta)=-\sum_{k=1}^K \sum_{\xx_n \in S_k} y_n \bx_n^T\bw_k I_{\{y_n \bx_n^T\bw_k<0\}}
\end{equation}
For a fixed $k$, $-\sum_{\xx_n \in S_k} y_n \bx_n^T\bw_k I_{\{y_n \bx_n^T\bw_k<0\}}$ is same as the Perceptron criterion function and we can
find $\bw_k$ to optimize this by using Perceptron algorithm. However, in $E_P(\Theta)$ defined by (\ref{error2}), the sets $S_k$ themselves are function
of the set of parameters $\Theta=\{\bw_1,\ldots,\bw_K\}$. Hence we can not directly minimize $E_P(\Theta)$ given by (\ref{error2}) using standard
gradient descent.

To minimize the Polyceptron criterion we adopt an alternating minimization scheme in the following way.
Let after $c^{th}$ iteration, the parameter set be $\Theta^c$. Keeping $\Theta^c$ fixed we calculate the sets
$S_k^c=\{\xx_n|\bx_n^T\bw_k^c \leq \bx_n^T\bw_j^c,~\forall j \neq k\}$. Now we keep these sets $S_k^c$ fixed.
Thus the Polyceptron criterion after $c^{th}$ iteration becomes
\begin{eqnarray}
\nonumber E_P^c(\Theta)=-\sum_{k=1}^K \sum_{\xx_n \in S_k^c} y_n \bw_k^T\bx_n I_{\{y_n\bx_n^T\bw_k<0\}}=\sum_{k=1}^K f_k^c(\bw_k)
\end{eqnarray}
where $f_k^c(\bw_k)=-\sum_{\xx_n \in S_k^c} y_n \bw_k^T\bx_n I_{\{y_n\bx_n^T\bw_k<0\}}$. Superscript $c$ is used to emphasize that the Polyceptron
criterion is evaluated by fixing the sets $S_k^c,~k=1\ldots K$.
Thus $E_P^c(\Theta)$ becomes a sum of $k$ functions $f_k^c(\bw_k)$ in such a way that $f_k^c(\bw_k)$
depends only on $\bw_k$ and it does not vary with the other $\bw_j,~\forall j\neq k$.

Now minimizing $E_P^c(\Theta)$ with respect to $\Theta$ boils down to minimizing each of $f_k^c(\bw_k)$ with respect to $\bw_k$.
For every $k\in\{1,\ldots, K\}$, a new weight vector $\bw_k^{c+1}$ is found using gradient descent update as follows.
\begin{eqnarray}
\nonumber \bw_k^{c+1} = \bw_k^c - \eta \frac{\partial E_P^c}{\partial \bw_k}= \bw_k^c + \eta \sum_{\xx_n \in S_k^c} t_n \bx_n
\end{eqnarray}
where $\eta$ is the step size. Here we have given only one iteration of gradient descent.
We may not minimize $f_k^c(\bw_k)$ exactly, so we may run a few steps of gradient descent.
We assume that $\eta$ is sufficiently small and hence the new weight vector $\bw_k^{c+1}$ is such that $f_k^c(\bw_k^{c+1})<f_k^c(\bw_k^c)$.
Then we calculate $S_k^{c+1}$ and so on.

To summarize, batch Polyceptron is an alternating minimization algorithm to
minimize the Polyceptron criterion. This algorithm first finds the sets $S_k^c,~k=1\ldots K$ for iteration $c$ and
then for each $k\in \{1,\ldots,K\}$ it learns a linear classifier by minimizing $f^c_k(\bw_k)$.
The minimization algorithm would be essentially same as the batch version of Perceptron algorithm.
We keep on repeating these two steps until there is no significant changes in
the weight vectors. Thus if the sum of the norms of gradients of Polyceptron criterion with respect to different weight vectors is less than a threshold, say $\gamma>0$, then the algorithm
stops updating the weight vectors.
The batch Polyceptron algorithm is described in Algorithm~\ref{algo1}.

\begin{algorithm}
\caption{Batch Polyceptron}
\label{algo1}
\KwIn{Training dataset $\{(\xx_1,y_1),\ldots,(\xx_N,y_N)\}$, $K$, $\eta$, $\gamma$}
\KwOut{$\{\bw_1,\ldots, \bw_K\}$}
\Begin
{
\textbf{Initialize} $\bw^0_1,\ldots,\bw^0_K$ and $S_1^0,\ldots,S_K^0$, set $c=0$\;
\While{$\sum_{k=1}^K||\sum_{\xx_n \in S_k^c} y_n \xx_n||<\gamma$}{
      $c\leftarrow c+1$\;
      \For{$k\leftarrow 1$ \KwTo $K$}{
            $\bw_k^c \leftarrow \bw_k^{c-1} + \eta \sum_{\xx_n \in S_k^{c-1}} y_n \bx_n$\;
            $S_k^c = \{\xx_n \in S| k=\amin_j \bx_n^T\bw_j^c\}$\;
      }
}
\textbf{return} $\bw_1,\ldots, \bw_K$\;
}
\end{algorithm}

\subsection{Online Polyceptron}
In the online Polyceptron algorithm, the examples are presented in a sequence.
At every iteration the algorithm updates
the weight vectors based on a single example presented to the algorithm at that iteration.
The online algorithm works in the following way.
The example $\xx_c$ at $c^{th}$ iteration is checked to see whether it is classified correctly
using the present set of parameters $\Theta^c$. If the example is correctly classified
then all the weight vectors are
kept same as earlier. But if $\xx_c$ is misclassified and $r={\arg\min}_{k\in \{1,\ldots,K\}}(\bw_k^c)^T\bx_c$, then only $\bw_r$
is updated in the following way.
\begin{eqnarray}\label{eq:online_update}
  \bw^{c+1}_r&=& \bw^c_r+ y_c\bx_c
\end{eqnarray}
Here we have set the step size as `1'. In general, any other appropriate step size can also be chosen.
The complete online Polyceptron algorithm is described in Algorithm~\ref{algo2}.
As is easy to see, the online Polyceptron can be thought of as an extension of Perceptron algorithm to the polyhedral set learning problem.

In the online Polyceptron algorithm, we see that the contribution to the error from $\xx_c$ will be reduced
because we have
\begin{equation}
-y_c(\bw_r^{c+1})^T\bx_c=-y_c(\bw_r^c)^T \bx_c - y_c^2 \bx_c^T \bx_c < -y_c(\bw_r^c)^T\bx_c \nonumber
\end{equation}
Although the contribution to the error due to $\xx_c$ is reduced, this does not
mean that the error contribution of other misclassified examples to the error is also reduced.

As is easy to see, online algorithm is very similar to the standard Perceptron algorithm. The original Perceptron algorithm converges in finite number of iterations if the training set is linearly separable. However, this does not imply that the Polyceptron algorithm
would converge if the data is polyhedrally separable. The reason for this is as follows: when $\xx_c$ is misclassified, we are using
it to update the weight vector $\bw_r$, where $r$ is chosen $r={\arg\min}_k y_c\xx_c^T\bw_k^c$. While this may be a good heuristic to decide which hyperplane
should take care of $\xx_c$, we have no knowledge of this. This is the same credit assignment problem that we explained earlier.
At present we have no proof of convergence of the online Polyceptron algorithm. However, given the empirical results presented in the next section, we
feel that this algorithm should have some interesting convergence properties.
\begin{algorithm}
\caption{Online Polyceptron}
\label{algo2}
\KwIn{$K$, A sequence of examples $(\xx_1,y_1),(\xx_2,y_2),\ldots(\xx_T,y_T)$}
\KwOut{$\bw_1,\ldots,\bw_K$}
\Begin
{
\textbf{Initialize: }$\bw^0_1,\ldots,\bw^0_K$\\
\For{$c\leftarrow 1$ \KwTo $T$}{
	    Get a new example $(\xx_c,y_c)$. Let $r=\amin_k (\bw^{c-1}_k)^T\bx_c$\;
	    Predict $\hat{y}_c=\mbox{sign}((\bw^{c-1}_r)^T\bx_c)$\;
	    \uIf{$\hat{y}_c \neq y_c$}{
		  $\bw^c_r \leftarrow \bw^{c-1}_r+\delta_{k,r}y_c\bx_c$\;		
	    }
	    \Else{
		  $\bw^c_k \leftarrow \bw^{c-1}_k,~k=1,\ldots,K$\;
	    }
      }
\textbf{return} $\bw_1,\ldots, \bw_K$\;
}
\end{algorithm}
\section{Experiments}\label{Sec:Experiments}
To test the effectiveness of Polyceptron algorithm, we test its performance on several synthetic and real world datasets. We compare our approach with OC1 \cite{Murthy1994} which is generic top down oblique decision tree algorithm.
We compare our approach with a constrained optimization based approach for learning polyhedral classifier discussed in \cite{Astorino}. This  approach successively solves linear programs. We call it PC-SLP (Polyhedral Classifier-Successive Linear Program) approach. We also compare our approach with a polyhedral learning algorithm called SPLA1 \cite{Naresh10}.
 Since the objective here is to explicitly learn the hyperplanes that define the polyhedral set, we feel that comparisons
with other general PR techniques (e.g., SVM) are not relevant.
\subsection{Dataset Description} We generate two synthetic polyhedrally separable datasets in different dimensions which are described below,
\begin{enumerate}
\item \textbf{Dataset~1 - 10 Dimensional Polyhedral Set: }1000 points are sampled uniformly from $[-1~~1]^{10}$. A polyhedral set is formed by
intersection of following three halfspaces.
\begin{enumerate}
 \item $x_1+x_2+x_3+x_4+x_5+x_6+x_7+x_8+x_9+x_{10}+1\geq0$
\item $x_1-x_2+x_3-x_4+x_5-x_6+x_7-x_8+x_9-x_{10}+1\geq0$
\item  $x_1+x_3+x_5+x_7+x_9+0.5\geq0$
\end{enumerate}
Points falling inside the polyhedral set are labeled as positive examples and the points falling outside this polyhedral set are labeled as negative examples.
\item \textbf{Dataset~2 - 20 Dimensional Polyhedral Set: }1000 points are sampled uniformly from $[-1 ~~1]^{20}$.
A polyhedral set if formed by intersection of following four halfspaces.
\begin{enumerate}
\item $x_1+2x_2+3x_3+4x_4+5x_5+6x_6+7x_7+8x_8+8x_9+8x_{10}+20x_{11}+
8x_{12}+7x_{13}+6x_{14}+5x_{15}+4x_{16}+3x_{17}+2x_{18}+x_{19}+x_{20}+20\geq0$
\item $-x_1+2x_2-3x_3+4x_4-5x_5+6x_6-7x_7+8x_8-9x_9+15x_{10}-11x_{11}+
10x_{12}-9x_{13}+8x_{14}-7x_{15}+6x_{16}-5x_{17}+4x_{18}-3x_{19}+2x_{20}+15\geq0$
\item $x_1+x_3+x_5+x_7+2x_8+8x_{10}+2x_{12}+3x_{13}+3x_{15}+3x_{16}+4x_{18}+4x_{20}+8\geq0$
\item $ x_1-x_2+2x_5-2x_6+6x_9-3x_{10}+4x_{13}-4x_{14}+5x_{17}-5x_{18}+6\geq0$
\end{enumerate}
Points falling inside the polyhedral set are labeled as positive examples and the points falling outside this polyhedral set are labeled
as negative examples.
\end{enumerate}
We also test Polyceptron on two real world datasets downloaded from UCI ML repository \cite{Asuncion+Newman:2007} which are described
in Table~\ref{table:details}.
\begin{table}[h]
\begin{center}
 \begin{tabular}{|p{1.5in}|p{.7in}|p{.7in}|}
\hline
Data set                & Dimension &   \# Points   \\ \hline
Ionosphere              &    34     &      351      \\ \hline
Breast-Cancer           &    10     &      683      \\ \hline
\end{tabular}
\caption{\footnotesize{Details of datasets used from UCI ML repository}} \label{table:details}
\end{center}
\end{table}
\subsection{Experimental Setup} We implemented {\em Polyceptron} in MATLAB. In the batch Polyceptron there are two user
defined parameters, namely the step size $\eta$ and the stopping criterion $\gamma$. For our experiments, we fix $\eta=.1$ and $\gamma=50$ for all datasets. In the
online Polyceptron, we have one user defined parameter {\em \# passes},
which is number of times the whole data is passes through the algorithm.
Different values of {\em \# passes} are used for different datasets and are mentioned in Table~\ref{table:table1}.
We implemented SPLA1 \cite{Naresh10} in MATLAB. In SPLA1, the number of hyperplanes are fixed beforehand.
We used BFGS approach for the maximization steps in SPLA1.
For OC1 we have used the downloadable package available from Internet \cite{OC1}. We implemented PC-SLP approach also in MATLAB.
All the user defined parameters for different algorithms are found using ten fold cross validation results.
All the simulations were done on a PC (Core2duo, 2.3GHz, 2GB RAM).

\subsection{Experimental Results}
\begin{table}
\begin{center}
  \begin{tabular}{|p{.6in}|p{2in}|p{0.75in}|p{.8in}|p{.75in}|}
 \hline
 Data set    &   Method                           &      Accuracy                &         Time(sec.)    &        \# hyps           \\
 \hline
Dataset~1     & Batch Polyceptron ($\gamma=50$)    &       95.05$\pm$0.94  &  0.03$\pm$0.008    &       3                         \\
             & Online Polyceptron ({\em \# passes}=300)       &      89.08$\pm$0.91          &     1.55$\pm$0.01     &       3                         \\
             &  SPLA1  & \textbf{97.88}$\pm$0.31 & 0.25$\pm$0.01 & 3\\
             &  OC1                               &      77.53$\pm$1.74          &     6.65$\pm$0.87     &       22.01$\pm$5.52            \\
             &  PC-SLP                            &      71.26$\pm$5.46          &     27.70$\pm$10.07   &       3                         \\
\hline
Dataset~2     & Batch Polyceptron ($\gamma=50$)    &      \textbf{94.56}$\pm$0.90 &     0.23$\pm$0.53     &       4                         \\
             & Online Polyceptron ({\em \# passes}=400)       &      94.34$\pm$1.96          &     1.76$\pm$0.16     &       4                         \\
             &  SPLA1   & 92.7$\pm$0.99 &  0.35 &  4  \\
             &  OC1                               &      63.64$\pm$1.48          &     10.01$\pm$0.67    &       27.36$\pm$6.98            \\
             &  PC-SLP                            &      56.42$\pm$0.79          &     189.91$\pm$19.73  &       4                         \\
\hline
Ionosphere   & Batch Polyceptron ($\gamma=50$)    &       89.68$\pm$1.28   &  0.04$\pm$0.004    &       2                         \\
             & Online Polyceptron ({\em \# passes}=500)       &      81.15$\pm$2.66          &     0.91$\pm$0.03     &       2                         \\
             &  SPLA1 &  \textbf{90.71}$\pm$1.37&  0.18 &   2\\
             &  OC1                               &      86.49$\pm$2.08          &     2.4$\pm$0.11      &       8.99$\pm$3.36             \\
             &  PC-SLP                            &      78.77$\pm$3.96          &     45.31$\pm$35.66   &       2                         \\
\hline
Breast       & Batch Polyceptron ($\gamma=50$)    &       \textbf{98.52}$\pm$0.09         &     0.08$\pm$0.01     &       2                         \\
Cancer       & Online Polyceptron ({\em \# passes}=500)       &      91.93$\pm$3.36          &     1.72$\pm$0.01     &       2                         \\
             &  SPLA1 & 96.25$\pm$0.36 &  0.12 &  2 \\
             &  OC1                               &      94.89$\pm$0.81          &     1.52$\pm$0.13     &       5.82$\pm$0.95             \\
             &  PC-SLP                            &      83.87$\pm$1.42          &     22.86$\pm$1.07    &       2                         \\
\hline
\end{tabular}
\caption{Comparison Results} \label{table:table1}
\end{center}
\end{table}

We now discuss performance of Polyceptron in comparison with other approaches on different datasets. The results provided are based on 10
repetitions of 10-fold cross validation. We show average values and standard deviation (computed over 10 repetitions) of accuracy,
time taken and the number of hyperplanes learnt. Note that SPLA1, PC-SLP and Polyceptron
are specialized methods to learn polyhedral classifiers, so we
fix the number of hyperplanes required beforehand.
On the other hand, OC1 is a top-down greedy approach for learning generic classifiers and does not require to
fix the number of hyperplanes.
The results are presented in Table~\ref{table:table1}. We show results of both batch and online Polyceptron.
Table~\ref{table:table1} shows results obtained with SPLA1, OC1 and SLP also for comparisons.

We see that batch Polyceptron is always better than online Polyceptron in terms of time and accuracy. In the online Polyceptron at any iteration
only one weight vector is changed when the current example is being misclassified. After every iteration the Polyceptron algorithm may not be improving as far as the Polyceptron criterion is concerned.
Because of this, online Polyceptron takes more time to find appropriate weight vectors to form the polyhedral classifier. On the other hand, batch Polyceptron minimizes the Polyceptron criterion using an alternating minimization scheme. And we observe experimentally that the Polyceptron criterion is monotonically decreased after every iteration using batch Polyceptron.

\begin{figure}[h]
 \begin{tabular}{c}
  \includegraphics[scale=.47]{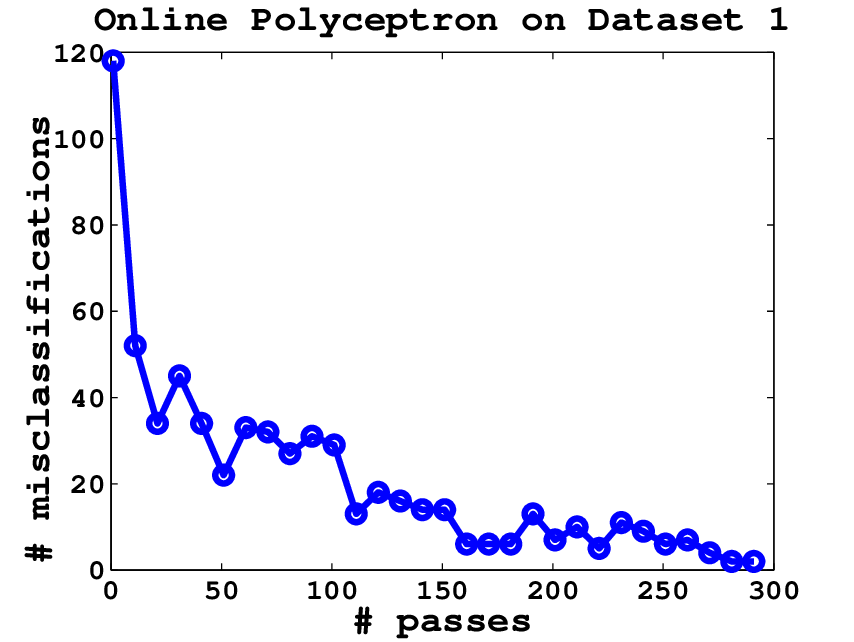}  \includegraphics[scale=.47]{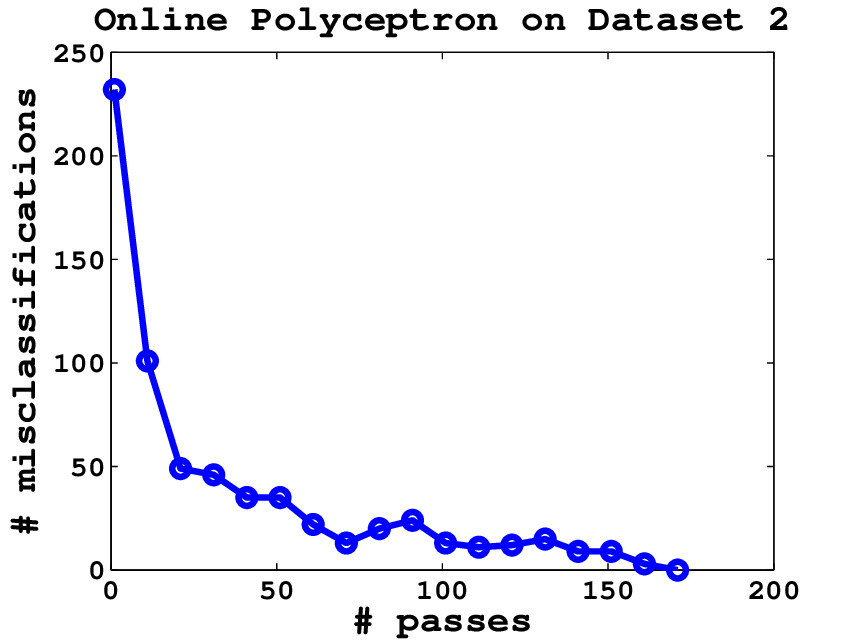}
 \end{tabular}
\begin{center}
\caption{Online Polyceptron on polyhedrally separable datasets. Converges in finite iterations.}
\label{online-poly}
\end{center}
\end{figure}

Fig.~\ref{online-poly} shows how \# misclassification goes down when online Polyceptron is run on Dataset~1 and Dataset~2 which are polyhedrally separable. We see that online Polyceptron converges in finite number of iterations. Which means, it learns a polyhedral classifier which correctly classifies all the training points. These experimental evidences raise an interesting question to ask
whether Polyceptron converges in finite iterations if the data is originally
polyhedrally separable.

We see that the batch Polyceptron performs comparable to SPLA1 in terms of accuracy.
Time wise, batch Polyceptron is atleast 1.5 times faster than SPLA1.

The results obtained with OC1 show that a generic decision tree algorithm is not good for learning polyhedral classifier.
Polyceptron learns the required polyhedral classifier with lesser number of hyperplanes compared to OC1 which is a generic decision tree algorithm. This happens because all other approaches here are model based approaches specially designed for polyhedral classifiers whereas OC1 is a greedy approach to learn general piecewise linear classifiers.
For synthetic datasets, we see that accuracies of both batch and online Polyceptron are greater than that of OC1 with a huge margin.
As the
dimension increases, the search problem for OC1 explodes combinatorially. As a consequence, performance of OC1 decreases as the
dimension is increased which is apparent from the results shown in Table~\ref{table:table1}. Also OC1, which is a general
decision tree algorithm gives a tree with a large number of hyperplanes.

For real word datasets, batch Polyceptron outperforms OC1 always. We see that for Breast Cancer dataset and Ionosphere dataset,
polyhedral classifiers learnt using batch Polyceptron give very high accuracy. This can be assumed that both these datasets are nearly
polyhedrally separable. In general, batch Polyceptron is much
faster than OC1.

Compared to PC-SLP \cite{Astorino}, Polyceptron approach always performs better in terms of both time and accuracy. As
discussed in Section~\ref{Sec:Introduction}, SLP which is a nonconvex constrained optimization based approach, has
to deal with credit assignment problem combinatorially which degrades its performance both computationally and time-wise. Polyceptron does not suffer from
such problem. The time taken by PC-SLP is much larger than any of the other algorithm.

Thus, in summary, the batch Polyceptron algorithm is a good method for learning polyhedral classifier and perform better than other available methods. The online Polyceptron algorithm is also
a fairly competitive method for the problem and it is an incremental algorithm. Given the obvious analogy with Perceptron, we feel that Polyceptron method is an interesting method for learning polyhedral classifiers.

\section{Conclusions}\label{Sec:Conclusion}
In this paper, we have proposed a new approach for learning polyhedral classifiers which we call {\em Polyceptron}.
We proposed Polyceptron criterion whose minimizer will give us the polyhedral classifier. To minimize Polyceptron criterion, we propose
online and batch version of Polyceptron algorithm.
Batch Polyceptron minimizes the Polyceptron criterion using an alternating minimization algorithm.
Online Polyceptron algorithm works like Perceptron algorithm as it updates the weight vectors only when there is a misclassification.
These are also interesting given the obvious relationship with the Perceptron algorithm.
We see that both the algorithm are very simple to understand and implement. We show experimentally that our approach efficiently finds
polyhedral classifiers when the data is actually polyhedrally separable. For real world datasets also our approach performs
better than any general decision tree method or specialize method for polyhedral sets.
Given the analogy between Perceptron and our Polyceptron algorithms, analyzing the convergence properties of Polyceptron algorithm would be an interesting problem for future work.
\bibliography{poly1}
\bibliographystyle{unsrt}
\end{document}